\pdfoutput=1

\documentclass[11pt]{article}

\usepackage{acl2023}

\usepackage{times}
\usepackage{latexsym}

\usepackage[T1]{fontenc}

\usepackage[utf8]{inputenc}

\usepackage{microtype}

\usepackage{inconsolata}

\usepackage{graphicx}

\usepackage{amsmath,amssymb}
\usepackage{hyperref}
\usepackage{multirow}
\usepackage{setspace}
\usepackage{multicol}
\usepackage{booktabs}
\usepackage{makecell}
\usepackage{xcolor}
\usepackage{array}
\usepackage{subcaption}
\usepackage[figuresright]{rotating}

\usepackage{placeins}

\title{Enhancing Accessibility of Medical Texts through Large Language Model-Driven Plain Language Adaptation}

\author{Ting-Wei Chang$^1$\quad \
        Hen-Hsen Huang$^2$\quad
        Hsin-Hsi Chen$^{1,3}$
    \vspace{5pt} \\
    \parbox{\textwidth}{
    \small
    \centering
    $^1$Department of Computer Science and Information Engineering, National Taiwan University,  Taiwan \linebreak
    $^2$Institute of Information Science, Academia Sinica,  Taiwan \linebreak
    $^3$AI Research Center (AINTU), National Taiwan University, Taiwan\quad
    }
    \vspace{2pt}\\
    \small
    \texttt{changtw@nlg.csie.ntu.edu.tw\quad hhhuang@iis.sinica.edu.tw}\quad
    \texttt{hhchen@ntu.edu.tw}
}

\begin{document}
\maketitle
\begin{abstract}

This paper addresses the challenge of making complex healthcare information more accessible through automated Plain Language Adaptation (PLA). PLA aims to simplify technical medical language, bridging a critical gap between the complexity of healthcare texts and patients’ reading comprehension. Recent advances in Large Language Models (LLMs), such as GPT and BART, have opened new possibilities for PLA, especially in zero-shot and few-shot learning contexts where task-specific data is limited. In this work, we leverage the capabilities of LLMs such as GPT-4o-mini, Gemini-1.5-pro, and LLaMA for text simplification. Additionally, we incorporate Mixture-of-Agents (MoA) techniques to enhance adaptability and robustness in PLA tasks. Key contributions include a comparative analysis of prompting strategies, finetuning with QLoRA on different LLMs, and the integration of MoA technique. Our findings demonstrate the effectiveness of LLM-driven PLA, showcasing its potential in making healthcare information more comprehensible while preserving essential content.
\end{abstract}

\section{Introduction}

Healthcare information is often presented in complex, technical language that can be challenging for the general public to understand. Yet, research highlights a persistent gap between the language used in medical documentation and the reading comprehension abilities of patients \cite{DBLP:journals/corr/abs-2012-12573, goldsack-etal-2022-making, luo-etal-2022-readability, Goldsack_2023}. Hence, PLA aims to bridge this gap by simplifying complex healthcare texts, making vital health information more accessible and promoting better health outcomes \cite{Goldsack_2023, mccray_promoting_2004}.

Developing automated solutions for PLA in medical terminology faces considerable challenges due to the complexity of extracting and simplifying specialized terms within biomedical texts. Recent methods have begun to incorporate semantic-sensitive approaches alongside word embedding techniques, such as the neighborhood context-based "Snowball" method, which has demonstrated promise in initial evaluations with expert-validated standards for term extraction \cite{9357263}. Another study attempts to provide comprehensive background explanations for key medical concepts in abstracts, relying on LLMs to accurately identify and interpret these concepts \cite{luo2024laypersonsguidebiomedicine}.

Recently, advancements in LLMs, such as T5, GPT and BART, have transformed various Natural Language Processing (NLP) tasks, offering potential solutions to PLA challenges \cite{li2024investigatinglargelanguagemodels, knappich2023boschaiplaba2023}. LLMs demonstrate promising abilities in lay summarization, particularly in zero-shot and few-shot learning contexts, where minimal task-specific data is available \cite{turbitt-etal-2023-mdc}. Other research addresses this problem by finetuning LLMs to improve their performance in enhancing the readability of biomedical texts \cite{li2024investigatinglargelanguagemodels, knappich2023boschaiplaba2023, sim-etal-2023-csiro, reddy-etal-2023-iitr}.

This paper compares and utilizes the latest models, including GPT-4o-mini, Gemini-1.5-pro, LLaMA, Gemma, and Mistral, along with techniques such as zero-shot, few-shot, QLoRA finetuning, and the advanced MoA methodology, which leverages the collective strengths of multiple LLMs \cite{wang2024mixtureofagentsenhanceslargelanguage}. MoA has demonstrated superior performance across various benchmarks, making it a compelling choice for complex tasks like PLA. The primary contributions of this work are outlined as follows:
\begin{enumerate}

    \item Comparison of prompting techniques: We conduct a thorough comparison of different prompting strategies, including zero-shot prompting, in-context learning (ICL), and ICL with semantic similarity \cite{liu2021makesgoodincontextexamples}. 
    \item Evaluation of finetuning approaches: We explore the effectiveness of finetuning advanced LLMs using QLoRA \cite{dettmers2023qloraefficientfinetuningquantized}, assessing how this technique impacts the models' ability to generate simplified medical language.
    \item Integration of MoA techniques: We utilize the MoA technique, effectively combining the unique capabilities of various LLMs to compare and analyze its impact on overall performance in PLA tasks.
    \item Automated evaluation and LLM judging: To enhance the robustness of our results, we implement an automated evaluation framework alongside LLM judging mechanisms. This dual approach allows for a comprehensive analysis of the outputs, ensuring that our findings are both reliable and actionable.
    \item We evaluate the new term replacement task introduced in PLABA 2024 using various prompt techniques.
\end{enumerate}

\section{Related Work}
In this section, we first introduce previous Plain Language Adaptation of Biomedical Abstracts (PLABA) work \cite{attal_dataset_2023}, then introduce biomedical text simplification, followed by lay summarization, and finally the efficient finetuning approach used in our work.
\subsection{PLABA}
The PLABA dataset addresses the challenge of simplifying complex biomedical literature in sentence level for general audiences. Despite the availability of health-related resources like MedlinePlus \footnote{https://medlineplus.gov/}, many scientific articles remain inaccessible to the public due to their specialized language. While various efforts have aimed to adapt technical terms for readability, creating manual plain language summaries for every medical article is impractical. Automated adaptation, supported by language models, has emerged as a promising solution, but requires high-quality, sentence-aligned datasets to train effective models. Existing datasets often suffer from imperfect alignments or lack sufficient scale, making them suboptimal for training and evaluation. PLABA fills this gap by providing 750 manually adapted abstracts from PubMed, offering sentence-level alignment for 7,643 pairs, each crafted by expert annotators. This dataset enables document- and sentence-level simplification and incorporates sentence-splitting to improve readability, making it a valuable gold standard for evaluating adaptation techniques in biomedical text simplification.

\subsection{Biomedical text simplification}
Language models like LLaMA-2 have been effectively applied to this task. For example, a LLaMA 2-based system achieved top performance in the previous PLABA shared task, focusing on simplifying complex biomedical text \cite{knappich2023boschaiplaba2023}. This approach highlights the difficulty in training models with high token overlap between input and output texts, which can limit the model's ability to perform substantive edits. To address this, sentence- and token-level loss weights were introduced, giving more emphasis to modified tokens, which led to simplifications closely aligned with human-generated adaptations, showing improved SARI and FKGL scores.

Another study explored various powerful LLMs, including encoder-decoder models (T5, SciFive, BART), GPT models (GPT-3.5, GPT-4), and control-token mechanisms within BART-based models, to simplify biomedical abstracts using the PLABA dataset \cite{li2024investigatinglargelanguagemodels}. Through domain-specific finetuning and prompt-based learning, these models were evaluated on both automated and human metrics. The BART-Large model with control token achieved the highest SARI score, while T5-Base scored best on BERTScore, balancing simplicity and meaning preservation. 

\subsection{Lay summarization}
The BioLaySumm shared tasks in 2023 and 2024 represent significant efforts in advancing lay summarization of biomedical research \cite{goldsack-etal-2023-biolaysumm, goldsack-etal-2024-overview}. These tasks, unlike PLABA tasks, focus on generating comprehensible summaries for non-expert audiences by using abstractive summarization techniques. In the 2023 task, models were trained to produce “lay summaries” that capture the essence of a full article while remaining accessible to general readers. Building on its success, the 2024 edition expanded participation and saw a trend towards innovative approaches, particularly with LLMs, reflecting the growing emphasis on this area.

\subsection{Efficient finetuning}
Efficient finetuning techniques like QLoRA are crucial for reducing computation costs and accelerating the finetuning process while retaining the full performance of 16-bit finetuning \cite{dettmers2023qloraefficientfinetuningquantized}. This is achieved by using a 4-bit quantized model and Low Rank Adapters (LORA) to backpropagate gradients without unfreezing the pretrained model weights \cite{hu2021loralowrankadaptationlarge}, leading to a substantial reduction in memory usage without sacrificing output quality. QLoRA's efficiency allows extensive experimentation across multiple model architectures, parameters, and instruction-following datasets, highlighting that high-quality, smaller datasets often outperform large, less-focused datasets in instruction finetuning. QLoRA’s open-source release, along with comprehensive analyses, provides a valuable framework for efficient, high-performance finetuning in natural language processing.
\subsection{MoA}
The MoA methodology represents an innovative approach in leveraging the combined expertise of multiple LLMs to enhance natural language understanding and generation tasks \cite{wang2024mixtureofagentsenhanceslargelanguage}. Unlike traditional single-model setups, MoA organizes LLMs into layers, with each agent in a given layer receiving input from the outputs of agents in the previous layer. This collaborative structure capitalizes on the "collaborativeness" phenomenon—where LLMs produce improved responses when they can build upon outputs from other models, even those of lower quality. MoA achieves state-of-the-art performance on several benchmarks and surpasses leading models like GPT-4 Omni. 

\section{Methodologies}
\begin{figure}[t]
  \centering
  \includegraphics[width=0.9\linewidth]{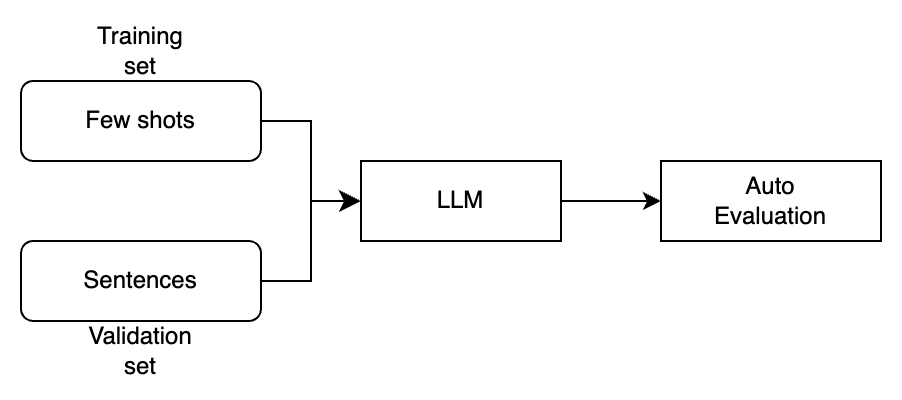}
  \caption{Pipeline of the PLA task in a few-shot setting. The provided annotated data is split into training and validation sets.}
  \label{fig:fig1}
\end{figure}

\begin{figure*}[t]
  \centering
  \includegraphics[width=0.9\linewidth]{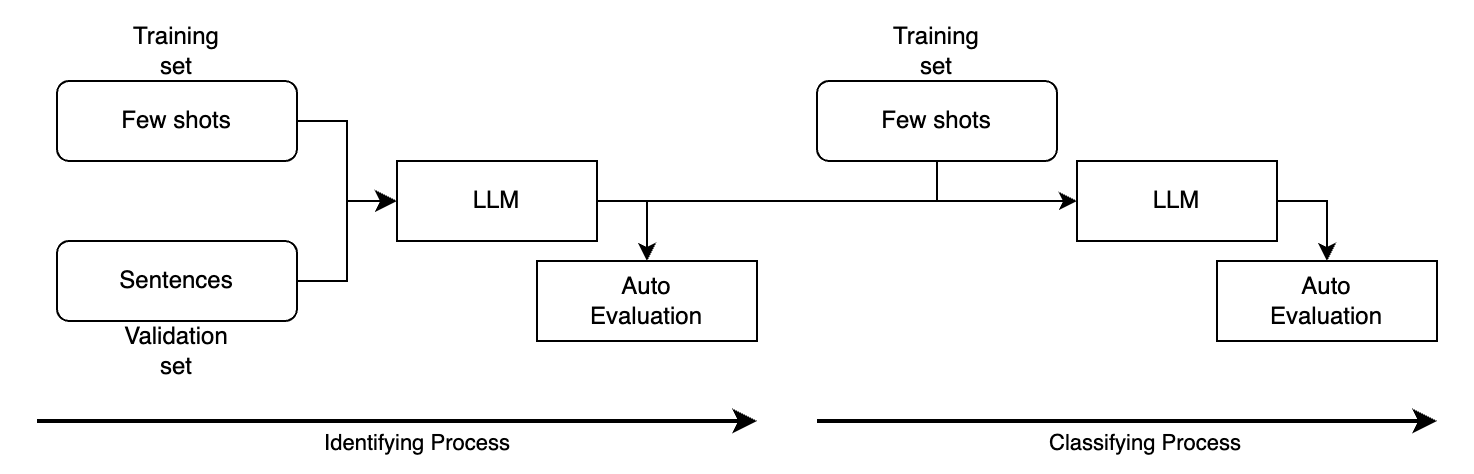}
  \caption{Pipeline of term replacement, containing Identifying process and Classifying process. The provided annotated data is split into training and validation sets.}
  \label{fig:fig2}
\end{figure*}

The overall framework of our experimental design is illustrated in Figures \ref{fig:fig1} and \ref{fig:fig2}, which outline the methodologies for the PLA task and the term replacement task.

The PLA Task aims to simplify each abstract while keeping sentences separate and ensuring accuracy and clarity for general readers. To accomplish this, we employ models using zero-shot prompting, random few-shot ICL, few-shot ICL enhanced by semantic similarity, and a zero-shot approach following QLoRA finetuning of LLMs on training data.

New to PLABA 2024, the Term Replacement Task focuses on identifying and simplifying technical terms within abstracts. Our approach starts by identifying complex terms in medical texts, then classifies the type of replacement needed, and finally generates lay language equivalents. This process is performed using few-shot ICL, with a focus on enhancing semantic similarity for more accurate replacements.

\subsection{Models}
Our experiments utilize a diverse set of advanced LLMs, each with unique capabilities suited to NLP tasks.
\subsubsection{Gemini}Gemini-1.0-pro, Gemini-1.5-flash, Gemini-1.5-pro are developed by Google, the Gemini models are multimodal, capable of handling image, audio, video, and text. Gemini-1.5 models, such as Pro and Flash, are particularly known for their efficiency in handling large contexts, enabling detailed recall over long text and multimedia inputs.
\subsubsection{GPT-4o-mini}A cost-efficient model from OpenAI, GPT-4o-mini is optimized for high performance on natural language tasks while lowering the costs and latency.
\subsubsection{LLaMA}LLaMA2 (7B), LLaMA3 (8B, 70B), LLaMA3.1 (8B, 70B) are developed by Meta, offer enhanced support for multilinguality, reasoning, and long-context processing. LLaMA 3.1 supports up to 128K tokens, allowing for complex understanding and generation tasks.

\subsubsection{Gemma 2}The Gemma 2 (2B, 9B, 27B) series from Google DeepMind is a lightweight yet capable family of models, scaling up to 27 billion parameters. These models leverage advanced techniques such as interleaving local-global attentions and grouped-query attention, along with knowledge distillation, enabling smaller models to achieve performance comparable to larger ones.

\subsubsection{Mistral}The Mistral 7B model uses grouped-query attention and sliding window attention, which significantly improves inference speed and allows handling of longer sequences, making it suitable for real-time applications. Mistral NeMo, developed in collaboration with NVIDIA, is a larger model designed to handle even more extensive tasks with a context window of up to 128K tokens. 

\subsection{QLoRA}
In our experiments, we applied QLoRA finetuning to optimize model efficiency and memory usage. The finetuning process employed a 4-bit quantized model to significantly reduce computational load. We set both the LoRA rank and LoRA alpha to 16, used a learning rate of 2e-4, kept the training batch size minimal at 2, and completed the finetuning in a single epoch.

\subsection{MoA}
We apply the MoA approach across different settings, utilizing Gemini-1.5-Flash to perform Aggregate-and-Synthesize while simplifying the process by employing only a single-layer MoA. Our approach builds upon the 5-shot setting from previous steps, using the following models as the adaptation models: Gemini-1.0-Pro, Gemini-1.5-Flash, Gemini-1.5-Pro, Gemma-2-27B, GPT-4o-Mini, Meta-Llama-3.1-8B, and Mistral-Nemo-Instruct-2407. Additionally, we explore the integration of finetuned models within the MoA framework by incorporating Gemma-2-27B, Meta-Llama-3.1-8B, and Mistral-Nemo-Instruct-2407, allowing us to assess the impact of finetuning on MoA’s performance in PLA tasks.

\subsection{Metrics for PLA task}
We apply automatic evaluation metrics to assess the performance of the PLA task, referencing the scoring methods used in PLABA2023 works and BioLaySumm \cite{li2024investigatinglargelanguagemodels, goldsack-etal-2024-overview, goldsack-etal-2023-biolaysumm}.
\begin{itemize}
    \item Relevance: BLEU \cite{Papineni2002BleuAM}, ROUGE \cite{lin-2004-rouge}, BERTScore \cite{zhang2020bertscoreevaluatingtextgeneration}, and SARI \cite{xu-etal-2016-optimizing}.
    \item Readability: Flesch-Kincaid Grade Level (FKGL), Dale-Chall Readability Score (DCRS), Coleman-Laiu Index (CLI).
    \item Factuality: AlignScore \cite{zha-etal-2023-alignscore}, SummaC \cite{laban2021summacrevisitingnlibasedmodels}.
    \item LLM Judge: Simplicity, Accuracy, Completeness, Brevity \footnote{https://bionlp.nlm.nih.gov/plaba2024/}.
    
\end{itemize}
Metrics like BLEU, ROUGE, BERTScore, and SARI are used to evaluate how closely the generated text aligns with reference summaries or simplified texts. BLEU and ROUGE measure the overlap of n-grams between generated and reference texts, with BLEU focusing on precision and ROUGE on recall. BERTScore uses contextual embeddings to evaluate semantic similarity, capturing meaning beyond surface-level token matches. SARI specifically assesses simplification by measuring edits made to the source text to match references, rewarding appropriate additions, deletions, and modifications, making it particularly suited for PLA tasks.

To gauge the accessibility of the text, readability metrics such as the Flesch-Kincaid Grade Level (FKGL), Dale-Chall Readability Score (DCRS), and Coleman-Liau Index (CLI) are employed. FKGL estimates the education level needed to understand the text, DCRS assesses readability based on word familiarity, and CLI evaluates readability through character and sentence length, with all three metrics indicating if the text is accessible to a general audience. In all three metrics, lower scores indicate greater simplicity and ease of comprehension.

For ensuring factual consistency, we apply AlignScore and SummaC. AlignScore is a versatile metric developed to handle factual inconsistency across diverse input-output pairs, leveraging a unified training approach that integrates data from multiple tasks, including NLI and QA, thus enhancing its generalizability. SummaC is designed specifically for summarization, using sentence-level analysis within documents to aggregate consistency scores. Both metrics provide a robust framework for detecting contradictions and ensuring the generated text remains factually aligned with the input information.

For LLM Judge, we utilized Gemini-1.5-flash and referenced prompts from previous work and made modifications to better suit our needs \cite{luo2024laypersonsguidebiomedicine}. The evaluation criteria were refined to focus on four main aspects: Simplicity, Accuracy, Completeness, and Brevity.

\subsection{Metrics for term replacement task}
After identifying difficult terms, we evaluate performance using the F1 score. For replacement type classification, we use the multilabel F1 score to measure accuracy. As for generating replacements, since no prior work offers a suitable reference, we did not conduct automatic evaluation for this aspect in this study.

\section{Experiments and Evaluations}
The PLABA dataset includes 750 biomedical abstracts that have been manually adapted into plain language by annotators, totaling 7,643 sentence pairs. In the 2024 PLABA track, additional tasks build upon this foundation. For the PLA task, 40 new consumer questions are introduced, each accompanied by 10 corresponding abstracts, resulting in a total of 400 test cases. In the Term Replacement Task, introduced exclusively in the PLABA 2024 track, 40 questions were selected from the original PLABA dataset. Each question is linked to 10 abstracts, with 10 questions provided as training data and the remaining 30 designated as test cases.

\subsection{Data preprocessing and splitting}
In the PLA task, we followed the data-splitting methodology outlined in the original PLABA dataset paper \cite{attal_dataset_2023}. The 921 abstracts were divided into approximately 85\% for the training set and 15\% for the validation set. Each topic was grouped and contained exclusively within either the training or test set to ensure unbiased evaluation. For the term replacement task, we split the provided annotated data into a 70/30 ratio.

\subsection{PLA task}
\subsubsection{Baseline from previous work}
We implemented the method from the original work \cite{attal_dataset_2023} and applied our evaluation method to establish a baseline comparison. The results are presented in Table \ref{tab:table_bs}.

\begin{table*}[]
\resizebox{\textwidth}{!}{%
\begin{tabular}{|c|c|c|c|c|c|c|c|c|c|c|c|}
\hline
Model          & BLEU           & SARI           & R-1            & R-2            & R-L            & BertS          & FKGL           & DCRS           & CLI            & AlignS         & SummaC         \\ \hline
T0PP           & 0.90           & 32.97          & 24.07          & 12.60          & 22.35          & 85.55          & 14.68          & 13.29          & 16.19          & 68.86          & 44.01          \\ \hline
Bart-base      & 24.37          & 40.13          & 59.39          & 30.41          & 56.71          & 89.21          & 12.89          & \textbf{10.88} & 13.85          & 61.53          & 45.71          \\ \hline
Bart-large-cnn & 24.31          & \textbf{40.29} & 59.32          & 31.09          & 56.72          & 89.25          & 12.72          & 11.20          & \textbf{13.83} & 70.42          & 50.57          \\ \hline
Bart-large     & 24.62          & 40.27          & 59.56          & 31.03          & 56.94          & 89.42          & \textbf{12.60} & 10.99          & 13.88          & 69.39          & 50.83          \\ \hline
Pegasus-large  & \textbf{25.83} & 37.70          & \textbf{60.47} & \textbf{32.51} & \textbf{57.88} & \textbf{89.70} & 12.93          & 11.00          & 14.40          & \textbf{74.80} & \textbf{55.89} \\ \hline
\end{tabular}%
}
\caption{Baseline evaluation scores for the models mentioned in the original paper. T0PP is not finetuned, while the others are finetuned models. In FKGL,DCRS,CLI, lower scores indicate greater simplicity and ease of comprehension. R = ROUGE, BertS = BertScore, AlignS =AlignScore}
\label{tab:table_bs}
\end{table*}

\subsubsection{Zero-shot vs ICL}
Table \ref{tab:table1} compares Gemini-1.0-pro, Gemini-1.5-flash, and GPT-4o-mini across different prompt methods. It shows that using few-shot ICL improves Relevance score, with ICL using semantic similarity performing slightly better than random few-shot selection. Different few-shot methods have minimal impact on the Readability and Factuality metrics.

\begin{table*}[h]
\centering
\resizebox{\textwidth}{!}{%
\begin{tabular}{|c|c|c|c|c|c|c|c|c|c|c|c|c|c|}
\hline
Model            & Type    & FewShot & BLEU           & SARI           & R-1            & R-2            & R-L            & BertS          & FKGL           & DCRS           & CLI            & AlignS         & SummaC         \\ \hline
Gemini-1.0-pro   & Zero    & 0       & 16.27          & 42.35          & 48.37          & 23.70          & 41.57          & 91.57          & 9.36           & 11.36          & 11.44          & 82.45          & 49.66          \\ \hline
Gemini-1.0-pro   & Closest & 1       & 23.73          & 42.59          & 54.63          & 32.25          & 49.23          & 92.35          & 11.27          & 12.01          & 12.86          & 85.13          & 57.91          \\ \hline
Gemini-1.0-pro   & Closest & 2       & 24.81          & 42.51          & 55.33          & 33.24          & 50.04          & 92.43          & 11.43          & 11.93          & 12.76          & \textbf{86.01} & 58.85          \\ \hline
Gemini-1.0-pro   & Closest & 3       & 26.23          & 42.52          & 56.41          & 34.45          & 51.37          & 92.57          & 11.67          & 12.08          & 12.94          & 85.99          & \textbf{60.26} \\ \hline
Gemini-1.0-pro   & Closest & 4       & 26.62          & 41.02          & 56.57          & 34.91          & 51.53          & 92.53          & 11.80          & 12.15          & 13.10          & 85.85          & 59.86          \\ \hline
Gemini-1.0-pro   & Closest & 5       & \textbf{27.26} & 41.45          & \textbf{57.08} & \textbf{35.53} & \textbf{52.19} & \textbf{92.59} & 11.99          & 12.17          & 13.21          & 85.50          & 60.08          \\ \hline
Gemini-1.5-flash & Zero    & 0       & 14.15          & 43.21          & 46.82          & 21.47          & 40.58          & 91.42          & \textbf{9.97}  & \textbf{10.88} & \textbf{11.16} & 71.05          & 36.92          \\ \hline
Gemini-1.5-flash & Random  & 1       & 16.90          & 44.00          & 49.03          & 24.12          & 42.83          & 91.72          & 10.39          & 11.21          & 11.60          & 74.76          & 41.01          \\ \hline
Gemini-1.5-flash & Random  & 2       & 18.17          & 44.11          & 50.19          & 25.30          & 44.10          & 91.84          & 10.34          & 11.29          & 11.78          & 74.47          & 41.16          \\ \hline
Gemini-1.5-flash & Random  & 3       & 19.31          & 45.07          & 51.51          & 26.72          & 45.42          & 92.02          & 10.49          & 11.36          & 11.84          & 75.18          & 42.94          \\ \hline
Gemini-1.5-flash & Random  & 4       & 20.18          & 44.77          & 52.34          & 28.15          & 46.68          & 92.06          & 10.84          & 11.43          & 11.98          & 74.13          & 43.09          \\ \hline
Gemini-1.5-flash & Random  & 5       & 21.24          & 45.51          & 53.06          & 28.97          & 47.36          & 92.13          & 11.05          & 11.52          & 12.08          & 74.29          & 43.71          \\ \hline
Gemini-1.5-flash & Closest & 1       & 16.97          & 44.21          & 49.11          & 24.33          & 42.86          & 91.70          & 10.31          & 11.10          & 11.49          & 73.99          & 40.87          \\ \hline
Gemini-1.5-flash & Closest & 2       & 18.85          & 45.30          & 50.55          & 26.12          & 44.41          & 91.82          & 10.44          & 11.19          & 11.57          & 74.08          & 41.78          \\ \hline
Gemini-1.5-flash & Closest & 3       & 19.81          & 45.59          & 51.77          & 27.47          & 45.88          & 92.00          & 10.67          & 11.25          & 11.72          & 72.51          & 41.88          \\ \hline
Gemini-1.5-flash & Closest & 4       & 21.00          & \textbf{46.11} & 52.98          & 28.95          & 47.07          & 92.08          & 10.93          & 11.37          & 11.93          & 73.09          & 42.75          \\ \hline
Gemini-1.5-flash & Closest & 5       & 21.37          & 46.04          & 53.57          & 29.70          & 47.97          & 92.08          & 11.11          & 11.43          & 11.98          & 71.91          & 42.47          \\ \hline
GPT-4o-mini      & Zero    & 0       & 12.37          & 41.95          & 45.07          & 18.89          & 38.00          & 91.31          & 11.22          & 11.30          & 11.52          & 76.24          & 38.09          \\ \hline
GPT-4o-mini      & Random  & 1       & 15.58          & 43.65          & 48.02          & 22.20          & 41.15          & 91.76          & 11.20          & 11.57          & 11.96          & 79.45          & 41.86          \\ \hline
GPT-4o-mini      & Random  & 2       & 16.62          & 44.03          & 49.02          & 23.21          & 42.48          & 91.90          & 11.43          & 11.74          & 12.19          & 79.89          & 42.53          \\ \hline
GPT-4o-mini      & Random  & 3       & 17.05          & 44.41          & 49.54          & 23.73          & 43.00          & 91.97          & 11.51          & 11.77          & 12.16          & 81.01          & 43.39          \\ \hline
GPT-4o-mini      & Random  & 4       & 17.36          & 44.67          & 50.33          & 24.19          & 43.45          & 92.00          & 11.64          & 11.81          & 12.21          & 80.66          & 43.73          \\ \hline
GPT-4o-mini      & Random  & 5       & 18.15          & 45.07          & 50.85          & 25.01          & 44.10          & 92.05          & 11.79          & 11.83          & 12.33          & 80.47          & 43.78          \\ \hline
GPT-4o-mini      & Closest & 1       & 16.10          & 44.17          & 48.57          & 22.67          & 41.77          & 91.82          & 11.43          & 11.61          & 11.91          & 79.63          & 42.19          \\ \hline
GPT-4o-mini      & Closest & 2       & 16.85          & 44.72          & 49.48          & 23.68          & 42.89          & 91.90          & 11.53          & 11.67          & 11.92          & 79.21          & 42.03          \\ \hline
GPT-4o-mini      & Closest & 3       & 17.45          & 45.07          & 50.34          & 24.40          & 43.74          & 91.99          & 11.51          & 11.67          & 11.99          & 80.11          & 42.55          \\ \hline
GPT-4o-mini      & Closest & 4       & 18.08          & 45.47          & 50.97          & 25.11          & 44.47          & 92.08          & 11.60          & 11.72          & 12.05          & 79.91          & 43.14          \\ \hline
GPT-4o-mini      & Closest & 5       & 18.44          & 45.62          & 51.34          & 25.70          & 44.91          & 92.12          & 11.67          & 11.74          & 12.11          & 79.62          & 42.95          \\ \hline
\end{tabular}%
}
\caption{Auto Evaluation of zero-shot versus different few shots type. "Type" refers to the few-shot ICL method: "Zero" indicates zero-shot, "Closest" represents few-shot ICL with closest semantic similarity, and "Random" indicates randomly selected few shots. "FewShot" indicates the number of few shots. In FKGL,DCRS,CLI, lower scores indicate greater simplicity and ease of comprehension. R = ROUGE, BertS = BertScore, AlignS = AlignScore}
\label{tab:table1}
\end{table*}

\subsubsection{Finetuning}
Table \ref{tab:table2} presents a comparison of various model series after finetuning. Results indicate a comprehensive improvement in Relevance and Factuality scores across all models. While Readability scores also increased. Using Gemini-1.5-flash as the LLM Judge, we observe that Simplicity remains relatively consistent, while Accuracy generally improves. Completeness and Brevity metrics show a uniform increase across models, highlighting the effectiveness of finetuning in enhancing these aspects.

\begin{table*}[]
\centering
\resizebox{\textwidth}{!}{%
\begin{tabular}{|c|c|c|c|c|c|c|c|c|c|c|c|c|c|c|c|c|}
\hline
Model                      & FT & BLEU           & SARI           & R-1            & R-2            & R-L            & BertS          & FKGL           & DCRS           & CLI            & AlignS         & SummaC         & Sim           & Acc           & Com           & Bre           \\ \hline
Llama-2-7b-chat            & X  & 7.36           & 40.58          & 35.92          & 14.77          & 30.79          & 88.92          & 10.62          & 10.36          & 11.63          & 46.00          & 31.68          & 8.22          & 8.46          & 8.09          & 6.32          \\ \hline
Llama-2-7b-chat            & O  & 24.79          & 39.42          & 51.67          & 32.87          & 48.04          & 91.21          & 12.16          & 12.83          & 15.16          & 58.20          & 48.32          & 8.47          & 8.32          & 8.13          & 8.21          \\ \hline
Llama-3-8b-Instruct        & X  & 11.44          & 41.33          & 44.88          & 18.62          & 36.80          & 91.01          & 11.20          & 11.28          & 12.11          & 73.58          & 38.50          & 8.83          & 9.06          & 8.69          & 7.88          \\ \hline
Llama-3-8b-Instruct        & O  & 33.77          & 35.31          & 61.67          & \textbf{43.04} & 58.44          & 93.14          & 13.17          & 12.79          & 14.47          & 88.82          & 70.85          & 8.70          & 9.46          & 9.48          & 9.12          \\ \hline
Llama-3-70b-Instruct       & X  & 10.79          & 40.77          & 42.24          & 16.45          & 35.02          & 90.90          & 10.82          & 10.77          & 10.94          & 72.52          & 34.22          & 8.86          & 9.01          & 8.60          & 7.92          \\ \hline
Llama-3-70b-Instruct       & O  & 33.38          & 37.84          & 61.81          & 42.82          & 58.48          & \textbf{93.23} & 13.00          & 12.67          & 14.29          & 86.63          & 67.84          & 8.77          & 9.38          & 9.37          & \textbf{9.13} \\ \hline
Llama-3.1-8B-Instruct      & X  & 9.41           & 38.96          & 39.69          & 14.80          & 32.82          & 90.41          & 9.76           & 10.53          & 10.87          & 71.03          & 34.43          & 8.84          & 8.59          & 7.99          & 8.03          \\ \hline
Llama-3.1-8B-Instruct      & O  & 31.75          & 40.57          & 60.19          & 40.21          & 56.40          & 92.95          & 12.50          & 12.41          & 13.75          & 80.79          & 61.38          & 8.79          & 9.24          & 9.15          & 8.82          \\ \hline
Llama-3.1-70B-Instruct     & X  & 13.25          & 42.81          & 45.05          & 19.97          & 38.18          & 91.18          & 10.26          & 10.52          & 10.40          & 73.52          & 38.17          & 8.82          & 9.03          & 8.63          & 7.95          \\ \hline
Llama-3.1-70B-Instruct     & O  & 31.85          & 41.39          & 60.87          & 41.17          & 57.14          & 93.07          & 12.51          & 12.41          & 13.80          & 81.61          & 61.12          & 8.81          & 9.20          & 9.09          & 8.90          \\ \hline
Mistral-7b-instruct-v0.3   & X  & 15.08          & 41.81          & 47.45          & 21.87          & 39.83          & 91.41          & 10.60          & 11.69          & 12.23          & 84.74          & 48.31          & 8.91          & 9.18          & 8.88          & 8.30          \\ \hline
Mistral-7b-instruct-v0.3   & O  & \textbf{33.82} & 33.75          & 60.71          & 42.27          & 57.61          & 92.88          & 13.22          & 12.78          & 14.52          & 87.08          & 72.15          & 8.44          & 9.44          & 9.42          & 8.83          \\ \hline
Mistral-Nemo-Instruct-2407 & X  & 13.09          & 41.09          & 44.36          & 19.55          & 38.14          & 91.35          & 8.94           & 11.41          & 11.34          & 81.14          & 44.52          & \textbf{8.92} & 9.10          & 8.66          & 8.29          \\ \hline
Mistral-Nemo-Instruct-2407 & O  & 31.83          & \textbf{42.83} & 60.37          & 40.75          & 56.85          & 92.98          & 12.17          & 12.23          & 13.48          & 75.02          & 56.36          & 8.90          & 9.04          & 8.91          & 8.84          \\ \hline
Gemma-2-2b-it              & X  & 1.97           & 40.30          & 34.71          & 13.04          & 29.13          & 89.05          & \textbf{8.59}  & \textbf{8.73}  & 10.36          & 60.25          & 31.08          & 8.18          & 8.59          & 8.31          & 6.39          \\ \hline
Gemma-2-2b-it              & O  & 33.03          & 34.17          & 61.09          & 42.25          & 57.85          & 93.07          & 13.06          & 12.77          & 14.44          & \textbf{90.46} & \textbf{73.02} & 8.60          & 9.45          & 9.48          & 9.08          \\ \hline
Gemma-2-9b-it              & X  & 8.73           & 38.89          & 38.73          & 14.14          & 32.07          & 90.32          & 9.17           & 10.09          & \textbf{10.15} & 65.73          & 31.64          & 8.80          & 8.93          & 8.46          & 7.67          \\ \hline
Gemma-2-9b-it              & O  & 33.51          & 35.35          & 61.66          & 42.82          & 58.41          & 93.12          & 13.09          & 12.76          & 14.46          & 88.52          & 70.92          & 8.68          & \textbf{9.47} & \textbf{9.48} & 9.08          \\ \hline
Gemma-2-27b-it             & X  & 10.29          & 39.75          & 41.41          & 15.75          & 34.19          & 90.76          & 8.89           & 10.59          & 10.52          & 71.31          & 34.35          & 8.87          & 9.05          & 8.59          & 7.95          \\ \hline
Gemma-2-27b-it             & O  & 33.78          & 36.80          & \textbf{61.84} & 42.90          & \textbf{58.60} & 93.20          & 12.99          & 12.66          & 14.34          & 86.15          & 68.68          & 8.66          & 9.46          & 9.40          & 9.00          \\ \hline
\end{tabular}%
}
\caption{Auto Evaluation of LLMs w/ or w/o finetuning. “FT” indicates whether finetuning is applied. 
R = ROUGE, BertS = BertScore, AlignS =AlignScore, Sim = Simplicity, Acc = Accuracy, Com = Completeness, Bre = Brevity}
\label{tab:table2}
\end{table*}

\subsubsection{MoA}
Table \ref{tab:tablemoa} presents a comparison of various model series after finetuning. Results indicate a comprehensive improvement in Relevance and Factuality scores across all models. While Readability scores also increased. Using Gemini-1.5-flash as the LLM Judge, we observe that Simplicity remains relatively consistent, while Accuracy generally improves. Completeness and Brevity metrics show a uniform increase across models, highlighting the effectiveness of finetuning in enhancing these aspects.
\begin{table*}[]
\centering
\resizebox{\textwidth}{!}{%
\begin{tabular}{|c|c|l|l|l|l|l|l|l|l|l|l|l|}
\hline
Model & FT & \multicolumn{1}{c|}{BLEU} & \multicolumn{1}{c|}{SARI} & \multicolumn{1}{c|}{R-1} & \multicolumn{1}{c|}{R-2} & \multicolumn{1}{c|}{R-L} & \multicolumn{1}{c|}{BertS} & \multicolumn{1}{c|}{FKGL} & \multicolumn{1}{c|}{DCRS} & \multicolumn{1}{c|}{CLI} & \multicolumn{1}{c|}{AlignS} & \multicolumn{1}{c|}{SummaC} \\ \hline
MoA   & X  & 8.49                      & 38.35                     & 40.43                    & 15.05                    & 33.0                     & 90.56                      & 9.07                      & 12.16                     & 12.74                    & 53.88                       & 32.64                       \\ \hline
MoA   & O  & 8.21                      & 38.05                     & 40.57                    & 15.14                    & 33.99                    & 90.53                      & 9.08                      & 12.27                     & 12.83                    & 54.84                       & 32.52                       \\ \hline
\end{tabular}%
}
\caption{Auto Evaluation of MoA w/ or w/o finetuned models. “FT” indicates whether finetund models are included. R = ROUGE, BertS = BertScore, AlignS =AlignScore }
\label{tab:tablemoa}
\end{table*}

\subsection{Term replacement task}
Table \ref{tab:table3} shows the evaluation of term replacement task. Compared to zero-shot, using few-shot ICL with the closest semantic similarity improves performance in both identifying difficult terms and classifying replacement types.

\begin{table}[]
\resizebox{\columnwidth}{!}{%
\begin{tabular}{|c|c|c|c|c|}
\hline
Model            & Identify shots & Idendify (F1)       & Classify shots & Classify (F1)       \\ \hline
Gemini-1.5-flash & 0              & 27.36          & 0              & 53.87          \\ \hline
Gemini-1.5-flash & 0              & 27.36          & 1              & 52.93          \\ \hline
Gemini-1.5-flash & 0              & 27.36          & 3              & 59.22          \\ \hline
Gemini-1.5-flash & 0              & 27.36          & 5              & 65.63          \\ \hline
Gemini-1.5-flash & 1              & 35.99          & 0              & 54.38          \\ \hline
Gemini-1.5-flash & 1              & 35.99          & \textbf{1}     & 54.45          \\ \hline
Gemini-1.5-flash & 1              & 35.99          & 3              & 61.22          \\ \hline
Gemini-1.5-flash & 1              & 35.99          & 5              & 66.75          \\ \hline
Gemini-1.5-flash & 3              & 35.69          & 0              & 56.77          \\ \hline
Gemini-1.5-flash & 3              & 35.69          & 1              & 56.19          \\ \hline
Gemini-1.5-flash & 3              & 35.69          & 3              & 63.75          \\ \hline
Gemini-1.5-flash & 3              & 35.69          & 5              & \textbf{68.52} \\ \hline
Gemini-1.5-flash & 5              & 38.98          & 0              & 55.98          \\ \hline
Gemini-1.5-flash & 5              & 38.98          & 1              & 55.46          \\ \hline
Gemini-1.5-flash & 5              & 38.98          & 3              & 62.33          \\ \hline
Gemini-1.5-flash & 5              & 38.98          & 5              & 67.12          \\ \hline
Gemini-1.5-pro   & 0              & 36.62          & 0              & 61.89          \\ \hline
Gemini-1.5-pro   & 0              & 36.62          & 1              & 60.99          \\ \hline
Gemini-1.5-pro   & 0              & 36.62          & 3              & 61.78          \\ \hline
Gemini-1.5-pro   & 0              & 36.62          & 5              & 64.86          \\ \hline
Gemini-1.5-pro   & 1              & 39.02          & 0              & 63.79          \\ \hline
Gemini-1.5-pro   & 1              & 39.02          & 1              & 62.22          \\ \hline
Gemini-1.5-pro   & 1              & 39.02          & 3              & 61.24          \\ \hline
Gemini-1.5-pro   & 1              & 39.02          & 5              & 64.90          \\ \hline
Gemini-1.5-pro   & 3              & 40.78          & 0              & 61.52          \\ \hline
Gemini-1.5-pro   & 3              & 40.78          & 1              & 59.00          \\ \hline
Gemini-1.5-pro   & 3              & 40.78          & 3              & 60.91          \\ \hline
Gemini-1.5-pro   & 3              & 40.78          & 5              & 63.99          \\ \hline
Gemini-1.5-pro   & 5              & \textbf{41.67} & 0              & 63.31          \\ \hline
Gemini-1.5-pro   & 5              & \textbf{41.67} & 1              & 62.30          \\ \hline
Gemini-1.5-pro   & 5              & \textbf{41.67} & 3              & 63.24          \\ \hline
Gemini-1.5-pro   & 5              & \textbf{41.67} & 5              & 66.96          \\ \hline
\end{tabular}%
}
\caption{F1 Scores for the Identification process and Multilabel F1 Scores for the Classification process in the Term Replacement Task. All few-shot examples were selected using ICL with closest semantic similarity}
\label{tab:table3}
\end{table}

\subsection{Shared Task Results on Test Dataset} 
\begin{figure*}[h]
    \centering
    \includegraphics[width=0.8\textwidth]{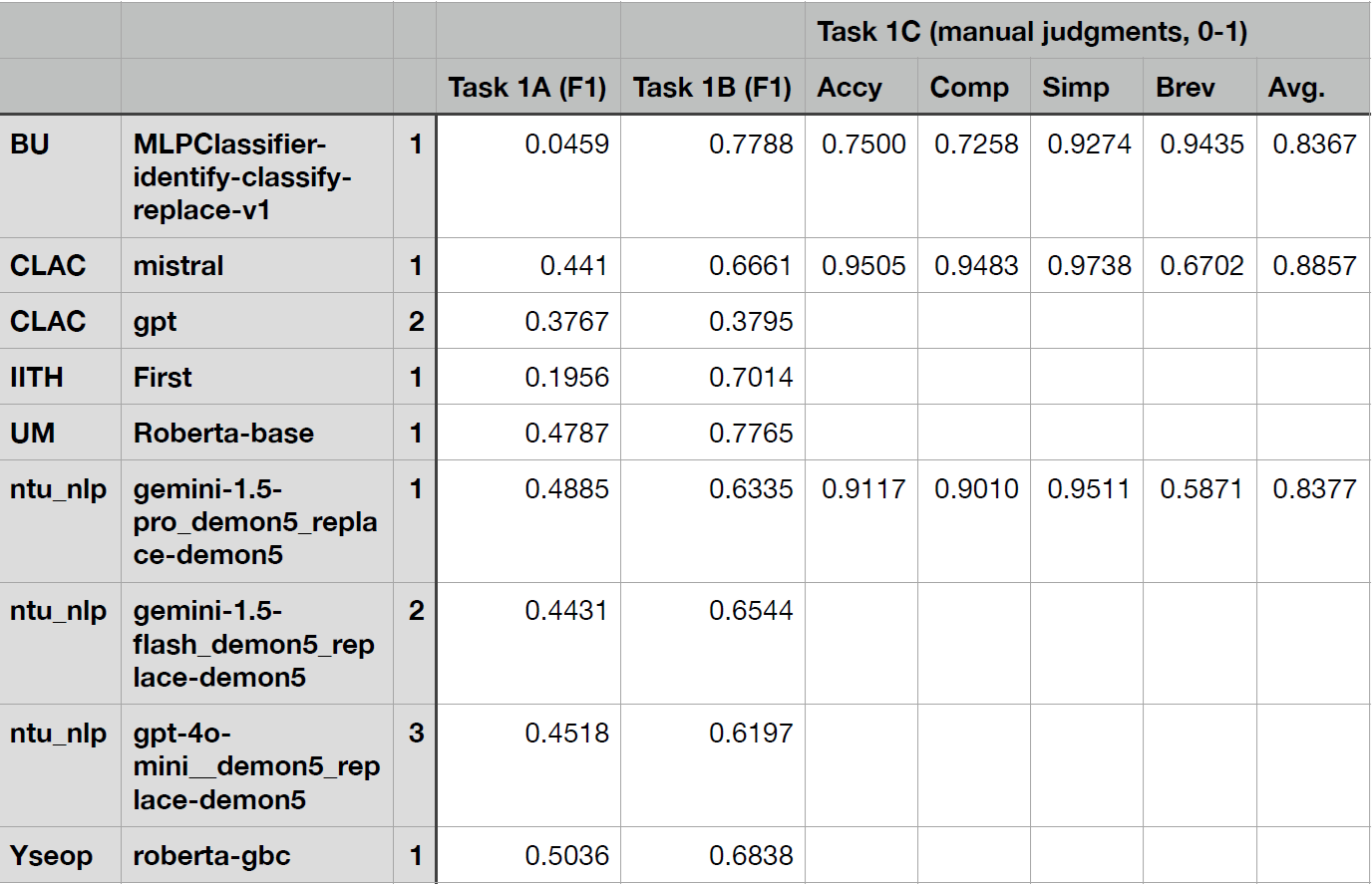} 
    \caption{Term Replacement Task Results \cite{ondov2024trec}: F1 scores for identifying difficult terms and classifying replacement types, with human evaluations for text generation. Evaluation metrics include Simplicity (Simp), Accuracy (Accy), Completeness (Comp), Brevity (Brev), and an overall Average (Avg.).} %
    \label{fig:task1}
\end{figure*}

Table \ref{tab:table_PLA} and Figure \ref{fig:task1} present the results of the test dataset as evaluated by the organizers \cite{ondov2024trec} for the PLA and term replacement tasks. The evaluation was conducted through human assessment, measuring four key criteria: Accuracy, Completeness, Simplicity, and Brevity, each scored on a scale of -1, 0, or 1. The final scores represent the average ratings across all evaluated samples. Some values remain unavailable due to the organizers' incomplete evaluation, attributed to a limited number of judges.

For the PLA task, we applied the MoA approach as described earlier. Our method ranked 4th when using finetuned models and 8th without finetuning.

In the term replacement task, all three models employed a 5-shot In-Context Learning (ICL) approach, selecting examples with the highest semantic similarity for both identification and classification. Our team ntu\_nlp achieved 2nd place in both the identification task and the overall average score for text generation, demonstrating its effectiveness in structured term adaptation.
\begin{table*}[h]
\centering
\resizebox{\textwidth}{!}{%
\begin{tabular}{|l|l|l|l|l|l|}
\hline
\multicolumn{1}{|c|}{Model}                      & Acc    & Com    & Sim    & Bre    & Avg    \\ \hline
GPT                                              & 0.9307 & 0.8118 & 0.9232 & 0.8755 & 0.8853 \\ \hline
LLaMA-8B-4bit-MedicalAbstract-seq-to-seq-v1      & 0.8831 & 0.8447 & 0.7792 & 0.5240 & 0.7578 \\ \hline
LLaMa\_3.1\_70B\_instruction\_2nd\_run           & 0.7440 & 0.7725 & 0.7044 & 0.5903 & 0.7028 \\ \hline
TREC2024\_SIB\_run1                              & 0.9374 & 0.8614 & 0.8363 & 0.6594 & 0.8236 \\ \hline
TREC2024\_SIB\_run3                              & 0.8170 & 0.7324 & 0.8756 & 0.7392 & 0.7910 \\ \hline
TREC2024\_SIB\_run4                              & 0.9433 & 0.8891 & 0.6965 & 0.6210 & 0.7875 \\ \hline
UAms-BART-Cochrane                               & 0.9772 & 0.9466 & 0.4881 & 0.6906 & 0.7756 \\ \hline
UAms-ConBART-Cochrane                            & 0.9546 & 0.9238 & 0.5653 & 0.6781 & 0.7804 \\ \hline
bart\_base\_ft                                   & 0.8424 & 0.7132 & 0.4500 & 0.4177 & 0.6058 \\ \hline
gpt-final                                        & 0.9537 & 0.8996 & 0.7850 & 0.3782 & 0.7541 \\ \hline
gpt35\_dspy                                      & 0.9117 & 0.8700 & 0.7654 & 0.6533 & 0.8001 \\ \hline
mistral-FINAL                                    & 0.8788 & 0.8567 & 0.7188 & 0.4122 & 0.7166 \\ \hline
mistral-fix                                      & 0.8818 & 0.8603 & 0.7118 & 0.4155 & 0.7174 \\ \hline
plaba\_um\_fhs\_sub1                             & 0.8985 & 0.8427 & 0.8421 & 0.7618 & 0.8363 \\ \hline
plaba\_um\_fhs\_sub2                             & 0.9447 & 0.8588 & 0.8909 & 0.7930 & 0.8718 \\ \hline
plaba\_um\_fhs\_sub3                             & 0.9504 & 0.8681 & 0.7949 & 0.6728 & 0.8215 \\ \hline
\textbf{task2\_moa\_tier1\_post (Ours MoA)}      & 0.8948 & 0.8463 & 0.8197 & 0.6272 & 0.7970 \\ \hline
\textbf{task2\_moa\_tier2\_post (Ours MoA w/ FT)} & 0.9307 & 0.8537 & 0.8642 & 0.6468 & 0.8238 \\ \hline
\end{tabular}%
}
\caption{PLA Task Results \cite{ondov2024trec}: Human evaluations for text generation. “FT” indicates whether finetund models are included. Evaluation metrics include Simplicity (Sim), Accuracy (Acc), Completeness (Com), Brevity (Bre), and an overall Average (Avg).}
\label{tab:table_PLA}
\end{table*}

\section{Discussion}
\subsection{Readability}
From the results in this study, models that were not finetuned showed greater variation in readability scores, likely due to differences in how each model interpreted the prompts. Most of these models tended to generate overly simplified outputs, potentially sacrificing content coherence and detail. Once finetuned, the readability of outputs across different models became more consistent, indicating that finetuning helped the models internalize the style and structure of expert-written plain language texts. This consistency suggests that finetuning enabled the models to adopt expert practices in plain language adaptation, improving the quality and readability of the outputs.

\subsection{Auto Evaluation}
Automated evaluation posed several challenges in this study. One of the primary issues was that standard readability metrics such as FKGL and DCRS are not designed to capture the nuances of plain language adaptation in medical texts. While these metrics provide a general measure of readability, they often fail to reflect the balance between simplifying content and maintaining its accuracy, which is critical in healthcare communication. 

\subsection{Other Dataset}
This study was limited to using the PLABA dataset for training and testing, which may not fully represent the range of medical and healthcare texts encountered in real-world applications. To more comprehensively evaluate the models' capabilities in text simplification, additional datasets specific to other medical domains or containing varied text structures would be beneficial. Access to a broader range of datasets could enable a more robust assessment of the model's ability to generalize across different contexts and further validate its adaptability to varied plain language adaptation tasks in healthcare.

\section{Conclusion}
In conclusion, this study demonstrates the potential of LLMs to simplify complex medical texts through PLA, leveraging few-shot ICL, finetuning, or the MoA approach. The few-shot approach with the closest semantic similarity improved the models' ability to generate relevant and readable outputs. finetuning also enhanced the effectiveness of PLA, enabling the models to adopt expert writing styles in plain language adaptation. While our results indicate that general-purpose models can effectively adapt to PLA tasks with suitable finetuning, the study also highlights limitations. All models tested were general-purpose LLMs, and none were pre-trained on medical datasets, which may have limited their ability to understand and perform on domain-specific tasks. Future work could explore the impact of models pre-trained on healthcare-specific data to further advance the effectiveness and accuracy of PLA in medical contexts.

\FloatBarrier
\bibliography{custom}
\bibliographystyle{acl_natbib}

\appendix

\end{document}